\documentclass[11pt,a4paper]{article}
\usepackage[hyperref]{acl2021}
\usepackage{times}
\usepackage{latexsym}
\usepackage{comment}

\usepackage{microtype}

\usepackage{graphicx} 
\usepackage{amsmath}
\usepackage{wrapfig,booktabs}
\usepackage{multirow}
\usepackage{algorithm,algorithmicx}
\usepackage[noend]{algpseudocode}
\usepackage{bbm}
\usepackage{wrapfig}
\usepackage{subfigure}

\newcommand{\cL}{\mathcal{L}}
\newcommand{\cB}{\mathcal{B}}
\newcommand{\cD}{\mathcal{D}}
\newcommand{\cM}{\mathcal{M}}

\aclfinalcopy 

\def\shownotes{0}  
\ifnum\shownotes=1
\newcommand{\authnote}[2]{[#1: #2]}
\else
\newcommand{\authnote}[2]{}
\fi

\title{Joint System-Wise Optimization \\for Pipeline Goal-Oriented Dialog System}

\author{Zichuan Lin\textsuperscript{\rm 1}, Jing Huang\textsuperscript{\rm 2}, Bowen Zhou\textsuperscript{\rm 2}, Xiaodong He\textsuperscript{\rm 2}, Tengyu Ma\textsuperscript{\rm 3} \\
\textsuperscript{\rm 1}Department of Computer Science and Technology, Tsinghua University\\
\textsuperscript{\rm 2}JD AI Research\\
\textsuperscript{\rm 3}Department of Computer Science, Stanford University\\
\texttt{lzcthu12@gmail.com,}\ \texttt{tengyuma@stanford.edu} \\ \texttt{\{jing.huang,bowen.zhou,xiaodong.he\}@jd.com} \\
}

\date{}

\begin{document}
\maketitle
\begin{abstract}


Recent work ~\cite{takanobu2020your} proposed the system-wise evaluation on dialog systems and found that improvement on individual components (e.g., NLU, policy) in prior work may not necessarily bring benefit to pipeline systems in system-wise evaluation. To improve the system-wise performance, in this paper, we propose new joint system-wise optimization techniques for the pipeline dialog system. First, we propose a new data augmentation approach which automates the labeling process for NLU training. Second, we propose a novel stochastic policy parameterization with Poisson distribution that enables better exploration and offers a principled way to compute policy gradient. Third, we propose a reward bonus to help policy explore successful dialogs. Our approaches outperform the competitive pipeline systems from \citet{takanobu2020your} by big margins of 12\% success rate in automatic system-wise evaluation and of 16\% success rate in human evaluation on the standard multi-domain benchmark dataset MultiWOZ, and also outperform the recent state-of-the-art end-to-end trained model from DSTC9~\cite{gunasekara2020overview}.


\end{abstract}

\section{Introduction}







\begin{figure*}[!t]
    \centering
    \includegraphics[width=0.90\textwidth]{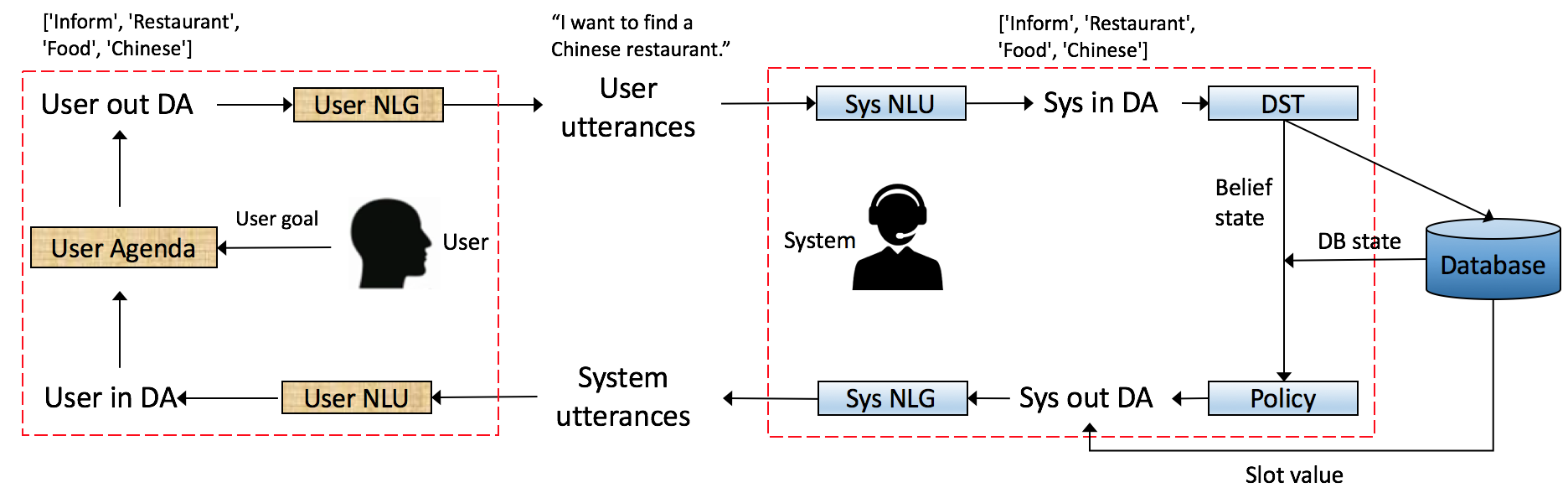}
    \caption{Workflow in a pipeline dialog system (by using an agenda-based user simulator). ``DA'' represents ``dialog acts''. First, user generates a user goal and user dialog acts, which follows by a user NLG to output user utterance. The system NLU takes in the user utterance and recovers dialog acts. The DST takes in the dialog acts, queries the database and returns belief state and DB state. The policy receives the state and outputs system out DA. Finally, the system NLG decodes the system out DA into system utterance and responses to the user.}
    \label{fig:pipeline}
\end{figure*}


Goal-oriented dialog systems have evolved from single domain to complex multi-domain tasks, with daily applications in customer support and personal assistants
~\cite{levin1997learning,stolcke2000dialogue,sarikaya2016overview,crook2016task,gao2018neural,zhang2020recent}. 
Existing approaches include 
(1) pipeline system, which typically consists of four modular components: \textit{Natural Language Understanding} (NLU)~\cite{goo2018slot,devlin2018bert,pentyala2019multi}, \textit{Dialog State Tracker} (DST)~\cite{xie2015recurrent,lee2016task}, \textit{Dialog Policy}~\cite{peng2017composite,takanobu2019guided}, and \textit{Natural Language Generation} (NLG)~\cite{wen2015semantically,balakrishnan2019constrained};
(2) end-to-end neural system with a learned model that takes in the conservation history and outputs the sentences~\cite{lei2018sequicity,zhang2020task,peng2020soloist,ham2020end}; 
and (3) some hybrid versions of (1) and (2)~\cite{zhao2019rethinking,lee2019sumbt}. Comparing with end-to-end systems, the modular structure makes the pipeline systems more interpretable. In this paper, we focus on the pipeline systems\footnote{We refer the readers to the right red box in Fig.~\ref{fig:pipeline} for an illustration of the pipeline system and the associated building blocks.}.

Recently \citet{takanobu2020your} proposed \textit{system-wise} evaluations on various configurations of goal-oriented dialog systems, which measured the success rate of the entire system to benchmark systematically. \citet{takanobu2020your} showed that improvement on individual components in prior work may not necessarily bring benefit to the pipeline systems in system-wise evaluation. 
More surprisingly, the best system (by a decent margin) in their paper turned out to be a simple pipeline system with a trained BERT-NLU~\cite{devlin2018bert}, rule-based DST, rule-based policy, and template-based NLG components. We refer to this system as the \textit{rule-based} system for simplicity. 
In this paper, we pose a question that has not been studied in prior work: what is the bottleneck that affects the pipeline system’s performance?

To answer this question, we propose joint system-wise optimization that trains components of the pipeline system jointly and synergistically to improve system-wise performance for pipeline systems. We train the NLU and policy together while fixing the DST and the NLG components as used in the {\em rule-based} system. This is because the rule-based DST and the template-based NLG both already have high performance. Therefore to simplify the joint system-wise optimization process, we focus on training the NLU and policy together with the fixed DST and NLG modules in the pipeline.

First, it is well known that BERT-NLU is data-hungry and requires a large amount of labeled training data~\cite{devlin2018bert}. However, labeled data is expensive and limited~\cite{zhang2020recent}. Existing labeling methods require humans to label intents and slots from utterances, which is quite time-consuming. In this paper, we design a novel automatic data labeling approach by leveraging the fact that the system NLU and the user NLG solve the inverse problems (as shown in Fig.~\ref{fig:pipeline}).
Given dialog acts that outputted by any user policies, we leverage a well-trained user NLG component to generate diverse utterances based on the dialog acts. 
Then, we train the system NLU by using the generated utterances as input and the dialog acts as labels. By training the system NLU component with this data augmentation technique, system NLU can achieve better performance and provide the system policy with correctly recognized intents/slots.

Second, we train the system policy to adapt to the output of its upstream components (i.e., system NLU). To encourage the exploration of policy, we propose a novel policy network with Poisson distribution to control the number of dialog acts. We also propose a reward bonus to help the policy reduce errors of the user NLU and explore successful dialog turns.
By training policy with enhanced exploration, the policy can explore more successful dialogs and provide the system NLU with more diverse training data.

We conduct extensive experiments on the multi-domain dialog benchmark dataset MultiWOZ 2.1~\cite{eric2019multiwoz} with agenda-based user simulator~\cite{schatzmann2007agenda} following the system-wise evaluation~\cite{takanobu2020your}. 
We show that our approaches significantly improve the learning efficiency of the pipeline dialog system and outperform the existing learning-based approaches as well as the rule-based system in both automatic evaluation and human evaluation. Our method also outperforms the best results (Team-1) in DSTC9~\footnote{In DSTC9, most teams adopt end-to-end neural systems based on GPT-2~\cite{radford2018improving} for NLU and NLG, while we take BERT for NLU and template-based method for NLG as used in \cite{takanobu2020your}} competition~\cite{gunasekara2020overview}, which is a concurrent work with ours. 

In summary, our contributions are:
\begin{itemize} 
	\itemsep -0.0em 
	\item[1.] We propose novel techniques to enable effective joint system-wise optimization for the pipeline dialog system.
	\item[2.] Our approaches outperform the competitive pipeline systems by big margins of 12\% success rate in automatic system-wise evaluation and of 16\% success rate in human evaluation on the multi-domain dataset MultiWOZ, and also outperform the recent state-of-the-art end-to-end trained model from DSTC9.
	\item[3.] Our ablation studies demonstrate that the bottleneck with the pipeline system comes from the NLU component.
\end{itemize}

\section{Related Work}

\subsection{Data Augmentation for NLU}
Similar to ours, \citet{liu2018dialogue} collected extra data to train the dialog system. However, their proposed approaches require access to human teaching, which is time-consuming and laborious.
\citet{liu2020robustness} proposed a model-agnostic toolkit LAUG to approximate natural perturbation and provided different data augmentation approaches to train NLU.
\citet{li2020textat} proposed adversarial token-level perturbation as data augmentation to improve the robustness of NLU.
\citet{wei2019eda} proposed four simple but powerful data augmentation operations to boost the performance of text classification tasks.
Different from the above work, our data augmentation leverages the inverse property of NLU and NLG, and is generated on-line with dialog conversations, which is more helpful for the NLU in goal-oriented dialog systems.

\subsection{Dialog Policy Learning}
Reinforcement Learning (RL) is commonly used to learn dialog policy, where users are modeled as a part of the environment and the policy is learned through interactions with users~\cite{zhao2016towards,liu2017iterative}.
\citet{peng2018deep} used pre-collected dialog acts as discrete actions and leveraged model-based reinforcement learning to train deep Q-networks~\cite{mnih2015human}. 

To further boost the learning efficiency of dialog systems, \citet{peng2017composite} formulated the task as options over Markov Decision Processes and use hierarchical RL to learn a dialog manager to operate at different option level. To leverage the human-human offline data, \citet{chen2017agent} addressed the problem of when and how to learn from the teacher's experiences. 
\citet{takanobu2019guided} proposed using a policy network with multinomial distribution to enlarge to exploration spaces and proposed an efficient reward estimation approach for efficient dialog policy learning. 
While these prior works mainly focused on improving dialog policy in component-wise evaluation, our method aims to boost the system-wise performance for the overall dialog system.

Our proposed stochastic policy parameterization is related to \citet{jhunjhunwala2020multi} in the sense that our dialog model samples multiple actions. \citet{jhunjhunwala2020multi} proposed to filter out invalid actions by rules and human interaction, while our stochastic policy parameterization enables better exploration and offers a principled way to compute policy gradient.



\subsection{End-to-End Neural Dialog System}
Our work is also related to end-to-end trained neural systems. An end-to-end trained model takes user utterances as input and directly outputs system responses in natural language, so it can be trained in a system-wise manner naturally. 
\citet{lei2018sequicity} proposed a holistic and extendable framework based on a single sequence-to-sequence (seq2seq) model~\cite{sutskever2014sequence} which can be optimized with supervised or reinforcement learning (RL) in end-to-end fashion. 
One drawback of end-to-end neural systems is that training the word-level policy with RL is very difficult due to the large action space. To mitigate this issue, \citet{zhao2019rethinking} proposed to learn policy networks in the latent action space. 
\citet{lee2019sumbt} proposed a universal and scalable belief tracker by jointly learning the NLU and DST modules, improving the flexibility of domain ontology configurations. 

Some recent work~\cite{hosseini2020simple,peng2020soloist,ham2020end} proposed a simple end-to-end neural system to predict belief state and sentence response jointly based on the strong auto-regressive models such as GPT-2~\cite{radford2019language}. Following this line, \citet{kulhanek2021augpt} and \citet{zhang2021hybrid} proposed improved pre-trained techniques and careful post-processing approaches to boost the performance of GPT-2 and achieved the best performance in DSTC9 competition~\cite{gunasekara2020overview}.
Despite the good performance and model simplicity, end-to-end neural systems are not as interpretable as the pipeline system.
In this paper, we focus on optimization for the pipeline dialog system and aim for system-wise improvement.

\section{Methodology}

To improve system-wise performance for the pipeline system, we propose joint system-wise optimization to train the components jointly and synergistically. As mentioned before, since rule-based DST and template-based NLG already achieve high performance, we focus on the joint training of NLU and policy.

On the one hand, training NLU requires a large amount of labeled data~\cite{devlin2018bert}. To ease the time-consuming labeling process, we design an automatic data augmentation approach by generating utterances from user NLG with dialog acts for training the system NLU 
(Section~\ref{method:NLU}).
On the other hand, the existing policy parameterization lacks mechanisms for adequate exploration. Therefore, we propose a novel stochastic policy parameterization as well as a reward bonus to encourage exploration (Section~\ref{method:RL}).

\subsection{Training NLU with Data Augmentation}
\label{method:NLU}

Let us denote $d$ as dialog act and $u$ as dialog utterance. We also denote the delexicalized dialog act as $\bar{d}$, which will be combined with the slot value from the database to result in a full dialog act. 
The NLU component $f_{\omega}$ maps an utterance $u$ to the corresponding dialog act $d$. We use the pre-trained BERT model as token encoder to train the NLU.
The NLG component maps the dialog act to an utterance. In this paper, we apply a template-based NLG implemented in Convlab-2~\cite{zhu2020convlab}.



We first observe that NLG and NLU are inverse processes of each other. 
Therefore, instead of labeling dialog acts from given utterances, our method generates utterances from given dialog acts. This allows us to adopt any well-trained user NLG models (or human feedback input if there is any) to provide diverse training data for system NLU.

Since human labeling is time-consuming, in this paper, we allow our system to interact with a user simulator (which is also a pipeline system) to collect data. The user simulator consists of BERT-NLU, rule-based DST, agenda-based policy~\cite{schatzmann2007agenda} and template-based NLG. 
During the conversation, the user simulator first outputs dialog acts $d$ and subsequently generates utterances $u$ by template-based NLG, which forms a new pair of training data ($u, d$) and provides training data augmentation for system NLU.

After collecting training data from the use simulator, we then mix the augmented data with the offline training data in MultiWOZ 2.1~\cite{eric2019multiwoz} together to fine-tune the system NLU.



We describe the NLU learning in more details below. The BERT-NLU consists of two sub-tasks: slot-value extraction ($f^{slot}$ ) and intent classification ($f^{intent}$). 
We use cross-entropy loss to train both the slot classification and the intent detection tasks as implemented in Convlab-2~\cite{zhu2020convlab}.
Let $\omega$ denote all the parameters and let 
$\cL_{slot}(\omega)$ and $\cL_{intent}(\omega)$ denote the two losses. The final loss function is the sum of two
\begin{equation} \label{eq:nluloss}
\cL_{slot}(\omega) + \cL_{intent}(\omega).
\end{equation}

To fine-tune the NLU, we first derive a rule-based algorithm to automatically convert the dialog acts $d$ to 
supervision signals for NLU (i.e., slot-values and intents). Then we use Eq.~\eqref{eq:nluloss} to fine-tune the NLU with the augmented training data.

Note that the template-based NLG in the user simulator can be replaced by any other well-trained NLG models (including humans). Moreover, using an ensemble of NLG models simultaneously can increase the diversity of training data, which helps enhance the robustness of system NLU. We leave it to future work.






\subsection{Exploration with Stochastic Policy Parameterization and Reward Bonus}  \label{method:RL}

The dialog policy generates the next system action conditioned on the dialog state. The dialog state consists of (1) belief state from DST; (2) DB state from database; (3) user actions at current turn; (4) system actions at the last turn. The dialog state is then represented as a binary vector of size $m$ ($m$ is the dimension of state) which serves as the input of policy. Given the input, the policy then outputs a distribution over all candidate delexicalized dialog acts and samples actions from that distribution.
Following common practice~\cite{takanobu2020your}, we use a sparse reward function: we give a positive reward $2L$ for a success dialog and a negative reward $-L$ for a failure one, where $L$ represents the number of dialog turns. we denote this reward function as $r_{origin}$.


To improve the performance of policy in system-wise evaluation, we train the policy using reinforcement learning in a system-wise manner by considering all the components (instead of assuming that they are perfect). Moreover, we propose two techniques to encourage the policy to explore successful dialogs.

\subsubsection{Stochastic Policy Parameterization}
Our starting point is the off-the-shelf RL algorithms PPO~\cite{schulman2017proximal}, which is one of the commonly used model-free algorithms in prior works in dialog~\cite{takanobu2019guided,takanobu2020your}. 
We improve the PPO algorithm based on the ConvLab-2 implementation~\cite{zhu2020convlab}.

First, we observe that existing policy parameterization maintains a separate Bernoulli distribution for each dialog act, and takes a simple threshold to output the set of dialog acts. There are several potential drawbacks: (1) the policy that collects data is deterministic, whereas the PPO algorithm requires a stochastic policy. Therefore there is a mismatch between the intended use of PPO and the actual implementation; (2) the determinism leads to insufficient exploration; (3) the parameterization does not utilize the mutual exclusiveness between different dialog acts.

\begin{figure}[t]
    \centering
    \includegraphics[width=0.45\textwidth]{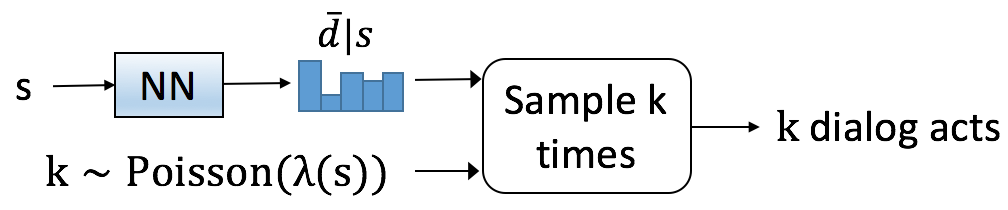}
    \caption{Policy network architecture.}
    \label{fig:policy}
\end{figure}

To overcome these drawbacks, we introduce a new parameterization of the stochastic policy that enables a principled way of using policy gradient algorithms as well as better exploration. 
As shown in Figure~\ref{fig:policy}, the policy network is given a state $s$ and outputs a sequence of delexicalized dialog acts (which will be combined with the slot value from the database to result in a full dialog act.) We first model the number of delexicalized dialog acts to output, denoted by $k$, by 
\begin{align}
    k \sim \textup{Poisson}(\lambda(s))
\end{align}
where $\lambda(s)$ is a function of the state parameterized by a neural net. 
We follow Convlab-2~\cite{zhu2020convlab} to use $209$ atomic actions. Then, we model the distribution of each delexicalized dialog act by a categorical distribution over the $209$ choices of delexicalized dialog act by 
\begin{align}
    \bar{d}|s \sim \textup{softmax}(g(s))
\end{align}
where $g(s)$ is a function of the state parameterized by a neural net that outputs a $209$-dimensional logit vector.
We assume all the $k$ dialog acts are independent with each other conditioned on $k$, resulting in the joint distribution of the actions $a = (
\bar{d}_1,\dots, \bar{d}_k)$:
\begin{align}
    \pi(a|s) = \frac{\lambda(s)^k}{k!} \cdot e^{-\lambda(s)} \cdot \prod_{1\le i\le k} p(\bar{d}_i|s).
\end{align}

We then compute the policy gradient update according to the PPO algorithm. During data collection, we directly sample from the stochastic policy instead of taking a deterministic threshold as done in the existing implementation~\cite{zhu2020convlab}. 

\subsubsection{Reward Bonus}
The second novel technique is motivated by the observation that when the user simulator is not perfect (e.g., the user NLU does not understand the system output utterances), the dialog system is not able to finish tasks successfully. 
We thus design a reward bonus to encourage the policy to select dialog acts whose translated utterances result in low errors for the user NLU (which is fixed without training) to reduce the errors from the user simulator.
We first measure the performance of user NLU by precision and recall at dialog act level and then use the F1 score as reward bonus $r_{bonus}$. 
Therefore, the final reward function is:
\begin{equation}
    r = r_{\textup{origin}} + \alpha \cdot r_{\textup{bonus}}
\end{equation}
where $\alpha$ is a hyper-parameter. 
The new reward function encourages the policy to lower the errors from the user simulator and explore successful dialogs during training.



\begin{algorithm}[htbp]
\caption{Joint System-Wise Optimization for Pipeline Systems (S-PPO)}
\label{alg}
\begin{algorithmic}[1]
    \State Given user NLU and NLG, rule-based DST, template-based NLG, data buffer $\cD$ and $\cM$, MultiWOZ dataset $\cB$. \label{code:init}
    \State Pre-train system NLU and user NLU on MultiWOZ dataset. Pre-train dialog policy $\pi_\theta$ (by assuming all the NLU and NLG are perfect, that is, passing along the dialog act between users and system directly). \label{code:pretrain}
    \For {each epoch}
        \State Trajectory data buffer $\cD \leftarrow \emptyset$, NLU data buffer $\cM \leftarrow \emptyset$
        \For {step $t=0,1,2,...,\textup{batch\_size}$}  \label{code:turn}
            \If {new session}
                \State Generate a user goal. \label{code:usergoal}
            \EndIf
            \State Receive user's utterance $u_{\textup{userout},t}$ with its corresponding dialog act $d_{\textup{userout},t}$.  \label{code:userutterance}
            \State Process user's utterance (with sampling from the policy) and send response to the user.\label{code:sys_response}
            \State Receive the immediate reward $r_{\textup{origin}}$ and  $r_{\textup{bonus}}$.
            \label{code:reward}
            \State Update the data buffers:
            $\cD \leftarrow \cD \cup \{(s_t,a_t,r_{\textup{origin}}+\alpha \cdot r_{\textup{bonus}})\}$, 
            $\cM \leftarrow \cM \cup \{(u_{\textup{userout},t}, d_{\textup{userout},t})\} $.
        \EndFor
        \State Train the dialog policy using PPO algorithm on collected data from $\cD$.  \label{code:updatepolicy} 
        \State Train system NLU on dataset $\cM$ and $\cB$ jointly to enforce consistency.  \label{code:updateNLU}
    \EndFor
\end{algorithmic}
\end{algorithm}

\subsection{Joint System-Wise Optimization}

Algorithm~\ref{alg} shows the pseudo code of our pipeline training process. We first initialize all components in the pipeline system (line \ref{code:init}), pre-train the NLU and dialog policy components (line \ref{code:pretrain}). For each new session, we first initialize a user's goal (line \ref{code:usergoal}). At each dialog turn, the system processes the user utterance, selects actions and responses to the users (line \ref{code:sys_response}). We then evaluate the current turn and calculate the original reward and reward bonus (line \ref{code:reward}). After collecting a batch of data, we train the dialog policy using PPO algorithm (line \ref{code:updatepolicy}) and train system NLU (line \ref{code:updateNLU}).

In our algorithm, the joint training of NLU and policy can benefit each other. On the one hand, a better NLU can provide more accurate input for policy; on the other hand, a good policy can explore better training data for NLU. By performing joint system-wise optimization for NLU and policy, the system-wise performance of pipeline systems can be improved.


%

\section{Experimental Results}

\subsection{Experimental setup}
We experiment on the common benchmark dataset MultiWOZ 2.1~\cite{eric2019multiwoz}, a multi-domain, multi-intent task-oriented dialog corpus that contains 7 domains, 13 intents, 25 slot types, 10,483 dialog sessions, and 71,544 dialog turns. We apply the agenda-based user simulator~\cite{schatzmann2007agenda}. 
The simulator initializes a user goal when the dialog starts, provides the agent with a simulated user response at each dialog turn and works at the dialog act level.

We compare our system with the following published baseline systems:
\begin{itemize}
    \item End-to-end trained Neural Systems: neural models that take in user utterances and output responses in natural language. We consider TSCP~\cite{lei2018sequicity}, DAMD~\cite{zhang2020task}, and SOLOIST+~\cite{zhang2021hybrid} which is a improved variant of SOLOIST~\cite{peng2020soloist} and achieved the best performance in DSTC9 competition~\cite{gunasekara2020overview}.
   \item Joint Systems: jointly-learning some components. We consider word-level DST SUMBT~\cite{lee2019sumbt} and word-level policy LaRL~\cite{zhao2019rethinking}.
   \item Modularized GDPL System (M-GDPL)~\cite{takanobu2019guided}: a dialog policy learning algorithm which uses inverse reinforcement learning for reward estimation.
   \item{Rule-based System (Rule)}~\cite{takanobu2020your}: achieves the state-of-the-art results in system-wise evaluation. It consists of a trained BERT-NLU, rule-based DST and policy, and template-based NLG components.
\end{itemize}

We also compare with Modularized PPO System (M-PPO)~\cite{takanobu2020your}, which is trained with PPO algorithm under the environment that assumes all the other components are exactly correct. Note that the only difference between M-PPO System and Rule-based System is about the policy. 
We refer to our proposed system as the {\bf Joint System-wise PPO System (S-PPO)}: we use our proposed joint system-wise optimization techniques to fine-tune the M-PPO system above. We also compare with a variant of S-PPO: we replace the learned policy in S-PPO with a rule-based policy --- we call it as Aug-Rule since it uses our data augmentation.

\begin{table*}[t]
	\centering
	\small
	\begin{tabular}{|c|cccc|}
		\hline
		& Turns & Info. & Match. & Succ.  \\
        \hline
        TSCP~\cite{lei2018sequicity} & 18.20 & 32.0 & 13.68 & 11.8 \\
        SUMBT~\cite{lee2019sumbt} & 13.71 & 44.0 & 46.44 & 27.8 \\
        LaRL~\cite{zhao2019rethinking} & 13.08 & 68.0 & 68.95 & 47.7 \\
        DAMD~\cite{zhang2020task} & 11.27 & 64.0 & 59.7 & 48.5 \\
        M-GDPL~\cite{takanobu2019guided} & 10.86 & 69.0 & 68.3 & 54.1 \\
        M-PPO~\cite{takanobu2020your} & 8.30 & 92.3 & 98.3 & 74.1 \\
        Rule~\cite{takanobu2020your} & \textbf{6.505} & 96.3 & 98.2 & 81.6 \\
        SOLOIST+~\cite{zhang2021hybrid} & 7.44 & 95.7 & 97.9 & 83.5 \\
        
        \hline
        Aug-Rule (ours) & 7.52 & 96.3 & 98.6 & 91.4 \\
        S-PPO (ours) & 7.21 & \textbf{98.4} & \textbf{99.6} & \textbf{93.8} \\
        \hline
	\end{tabular}
	\caption{System-wise automatic evaluation performance of dialog turns, inform recall, match rate, and success rate, for all baseline systems and our S-PPO system. S-PPO beats all baselines and achieves state-of-the-art results.}
	\label{table:main}
\end{table*}

\subsection{Training}

For all the RL-based methods, we use two hidden layer MLPs with 100 dimensions for the system policy and 50 dimensions for the value function. The action space of system policy is 209. The activation function is ReLU for MLPs. 
We follow \citet{takanobu2019guided} and pre-train the policy by simple imitation learning (with cross-entropy loss and a learning rate of 1e-5) until convergence on the state-action pairs on the MultiWOZ dataset~\cite{eric2019multiwoz}. 
We then use the pre-trained policy as initialization to interact with the user simulator and continue to train the policy using the PPO algorithm (M-PPO system). After that, we fine-tune M-PPO with our joint system-wise optimization techniques (S-PPO system). 
For PPO training, we use RMSprop optimizer with a learning rate of 1e-4 and a batch size of 2000. 
For NLU learning, we fine-tune all the parameters of NLU including the BERT model with AdamW~\cite{loshchilov2017decoupled} optimizer using a learning rate of 1e-4 and a batch size is 40 (including 32 augmented data and 8 offline MultiWOZ data).
For the reward bonus, we found that the coefficient $\alpha$ is not sensitive and we set $\alpha=10$.
For SOLOIST+~\cite{zhang2021hybrid}, we use the publicly available code and rerun their model in our test settings\footnote{Therefore result numbers are different from the original paper}.

\subsection{Evaluation Metric}
We use the number of \textit{dialog turns}, averaging over all dialog sessions to measure the efficiency of accomplishing a task. The system should help each user accomplish his/her goal within 20 turns, otherwise, the dialog is regarded as failed. We also utilize two other metrics: \textit{inform recall} and \textit{match rate} to estimate the task success. Both metrics are calculated based on the dialog act~\cite{stolcke2000dialogue}. The dialog act from the input and output of the agenda-based user simulator's policy will be used to calculate the two scores. \textit{Inform recall} evaluates whether all the information requests are fulfilled, and \textit{match rate} assesses whether the offered entity meets all the constraints specified in a user goal. The dialog is marked as successful if and only if both \textit{inform recall} and \textit{match rate} are 1. For each agent, the \textit{success rate} and other metrics are averaged on 1000 dialog sessions. 


\subsection{Main Results}
Table~\ref{table:main} shows the performance comparison with baselines. We compare all methods on the number of dialog turns, inform recall, match rate and success rate. 
Our method S-PPO outperforms end-to-end neural systems (TSCP, DAMD, SOLOIST), joint systems (SUMBT, LaRL) and other pipeline systems (M-GDPL, M-PPO, Rule) and achieves state-of-the-art success rate in system-wise evaluation. 
We discover that our method Aug-Rule also outperforms other baselines by a large margin. This demonstrates that our data augmentation approach can also significantly boost the performance of other pipeline systems.

Our proposed S-PPO system significantly outperforms the M-PPO system and beats the results of other systems by a large margin of 10\% in \textit{success rate}. 
S-PPO also achieves the best performance in \textit{inform recall}, \textit{match rate} and \textit{success rate}. This is because our techniques enable the system to understand and answer the user's requests, encourage the system to provide correct answers and thus have a higher success rate.
We also observe that the \textit{dialog turns} of S-PPO is lower than other learning-based systems and only slightly higher than that of the rule-based system.

\subsection{Human Evaluation Results}
We recruit 20 people to interact with dialog systems and collect their judgment on task success. Following the setting of DSTC competition~\cite{kim2019eighth}, we collect 100 dialogs for each system. For each dialog, we anonymize the system ID to reduce the user's bias.

\begin{table}[h]
	\centering
	\footnotesize
	\begin{tabular}{|cccc|}
	\hline
	
	Rule & SOLOIST+ & Aug-Rule & S-PPO \\
	\hline
	62\% & 77\% & \textbf{78\%} & \textbf{78\%} \\
	\hline
	\end{tabular}
	\caption{Success rate in human evaluation.} 
	\label{table:humaneval}
\end{table}

\begin{figure*}[t]
	\centering
	\includegraphics[width=0.23\textwidth]{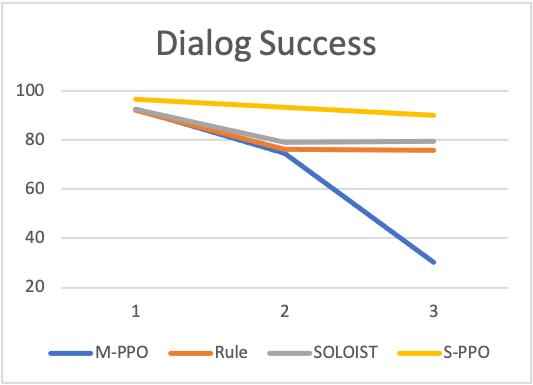}
	\includegraphics[width=0.23\textwidth]{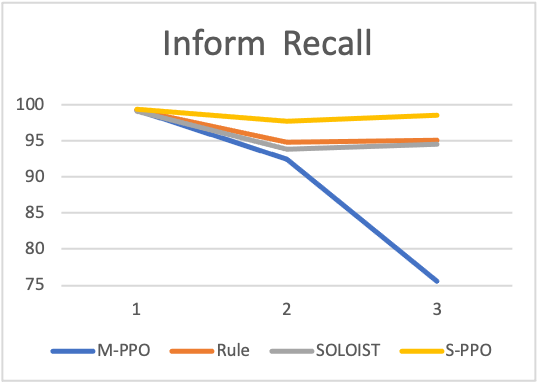}
	\includegraphics[width=0.23\textwidth]{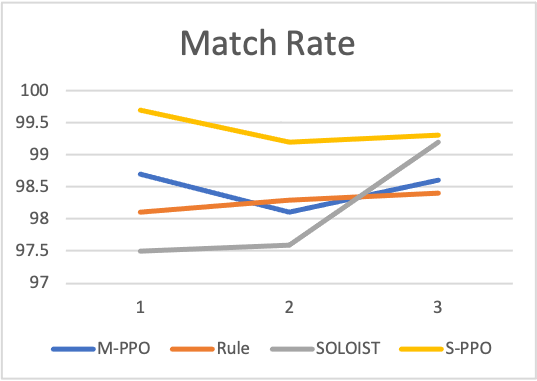}
	\caption{Performance (y-axis) with different number of domains (x-axis). S-PPO can handle complex tasks much better than baselines.}
	\label{fig:dom_number}
\end{figure*}

For each session, users are asked to mark whether the system completes the dialog. If a dialog is completed, users are also asked to provide all requested slot values for database query verification purposes. A dialog is successful if the fulfilled requested slots match the values in the database. We compare four systems which achieve the highest performance in automatic system-wise evaluation. The success rate of 62\% from the rule-based system is taken from~\cite{takanobu2020your}, while the success rate from our own human evaluation is lower than this number.

As shown in Table~\ref{table:humaneval}, S-PPO and Aug-Rule outperform other systems in human evaluation, which demonstrates the effectiveness of our approaches in real-world applications.
We find that S-PPO and Aug-Rule are significantly better than the rule-based system, which indicates that the systems trained with our data augmentation approach can understand human's utterances much better.
Our methods S-PPO and Aug-Rule also slightly outperform SOLOIST+, even though SOLOIST+ leverages a stronger GPT-2 model for language generation (NLG).
Table~\ref{case:sample} and \ref{case:sample-humaneval} in Appendix show the sampled dialog sessions of both automatic evaluation and human evaluation.

\subsection{Multi-domain tasks}
Figure~\ref{fig:dom_number} demonstrates how the performance varies with the number of domains in a task. S-PPO outperforms all baseline systems consistently when the number of domains increases from 1 to 3. We find that the performance of M-PPO drops dramatically when the number of domains increases. On the contrary, our proposed S-PPO system can scale well to the multi-domain tasks, and the performance of S-PPO only drops slightly when the number of domains increases. These results show that S-PPO can deal with complex tasks better than baselines.

\subsection{Ablation study}
In this section, we conduct ablation studies to investigate the contribution of the proposed techniques. Specifically, we test four variants: 1) `Vanilla': optimize the policy in a system-wise manner without using our approaches; 2) `Poiss': using stochastic policy parameterization with Poisson distribution; 3) `Aug': training system NLU with data augmentation; 4) `Bonus': train policy with reward bonus.

\begin{table}[h]
	\centering
	\small
	\begin{tabular}{|c|cccc|}
		\hline
		& Turn & Info. & Match. & Succ.  \\
		\hline
        Vanilla & 8.41 & 93.9 & 98.2 & 77.7 \\
        \hline
        Poiss & 8.23 & 94.6 & 98.8 & 80.3 \\
        \hline
        Poiss+Aug & 7.34 & 97.4 & 99.3 & 90.2 \\
        \hline
        Poiss+Aug+Bonus & 7.21 & 98.4 & 99.6 & 93.8 \\
        \hline
	\end{tabular}
	\caption{Ablative results in system-wise evaluation.} 
	\label{table:ablation}
\end{table}

The ablation results are shown in Table~\ref{table:ablation}. First, using our new parameterization of stochastic policy can improve the success rate by about 3\% comparing with the vanilla system. This is because the stochastic parameterization technique enables the control for the number of sampled dialog acts as well as better exploration. Second, training the system NLU with our proposed data augmentation can achieve the most performance gain (10\%). Third, using the reward bonus to train policy can improve the success rate by 3\%.
Our ablation results show that the main bottleneck with pipeline systems comes from the NLU component.

\begin{figure}[h]
	\centering
	\includegraphics[width=0.2\textwidth]{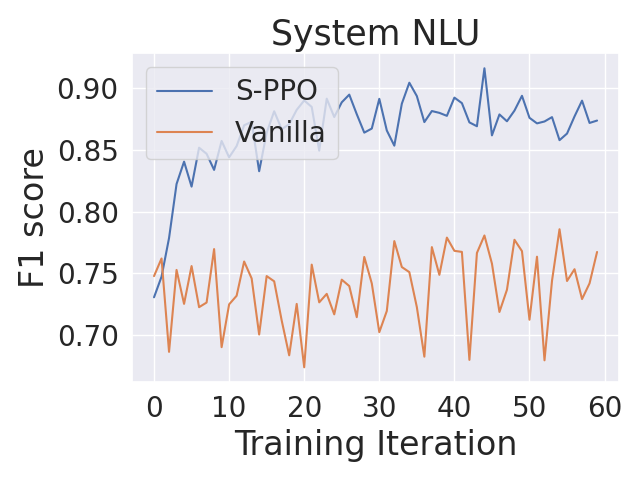}
	\includegraphics[width=0.2\textwidth]{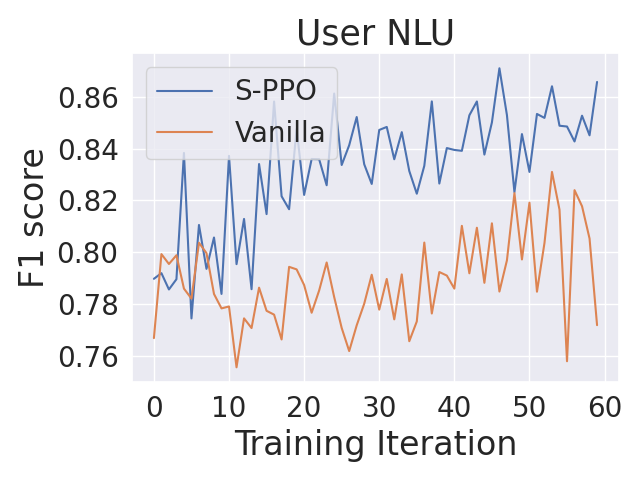}
	\vspace{-0.1in}
	\caption{Performance of system NLU and user NLU.}
	\label{fig:consistency}
	\vspace{-0.1in}
\end{figure}

To further show how our proposed techniques work, we provide analysis of intermediate results. Figure~\ref{fig:consistency} shows the F1 score of NLU during training. On the one hand, training \textit{system NLU} with data augmentation helps improve the performance. On the other hand, training policy with the reward bonus indirectly boosts the performance of \textit{user NLU}, even though we do not train the user NLU. 

\section{Conclusion}

In this paper, we propose novel joint system-wise optimization techniques for pipeline goal-oriented dialog systems. 
To mitigate the data sparsity problem of NLU, we propose to a novel data augmentation approach.
To enhance exploration of policy, we propose a novel stochastic policy parameterization with Poisson distribution as well as an additional reward bonus to encourage the policy to explore successful dialogs.
Our extensive experiments of automatic evaluation and human evaluation demonstrate that our approach outperforms prior works by a large margin and achieves state-of-the-art success rate in the system-wise evaluation on the 
multi-domain goal-oriented benchmark dataset MultiWOZ.
In the future, we plan to explore in the direction of: training the NLU and NLG components jointly; and applying our techniques to end-to-end neural dialog systems.


\bibliographystyle{acl_natbib}
\bibliography{anthology}

\begin{thebibliography}{45}
\expandafter\ifx\csname natexlab\endcsname\relax\def\natexlab#1{#1}\fi

\bibitem[{Balakrishnan et~al.(2019)Balakrishnan, Rao, Upasani, White, and
  Subba}]{balakrishnan2019constrained}
Anusha Balakrishnan, Jinfeng Rao, Kartikeya Upasani, Michael White, and Rajen
  Subba. 2019.
\newblock Constrained decoding for neural nlg from compositional
  representations in task-oriented dialogue.
\newblock \emph{arXiv preprint arXiv:1906.07220}.

\bibitem[{Chen et~al.(2017)Chen, Zhou, Chang, Yang, and Yu}]{chen2017agent}
Lu~Chen, Xiang Zhou, Cheng Chang, Runzhe Yang, and Kai Yu. 2017.
\newblock Agent-aware dropout dqn for safe and efficient on-line dialogue
  policy learning.
\newblock In \emph{Proceedings of the 2017 Conference on Empirical Methods in
  Natural Language Processing}, pages 2454--2464.

\bibitem[{Crook et~al.(2016)Crook, Marin, Agarwal, Aggarwal, Anastasakos,
  Bikkula, Boies, Celikyilmaz, Chandramohan, Feizollahi et~al.}]{crook2016task}
Paul~A Crook, Alex Marin, Vipul Agarwal, Khushboo Aggarwal, Tasos Anastasakos,
  Ravi Bikkula, Daniel Boies, Asli Celikyilmaz, Senthilkumar Chandramohan,
  Zhaleh Feizollahi, et~al. 2016.
\newblock Task completion platform: A self-serve multi-domain goal oriented
  dialogue platform.
\newblock In \emph{Proceedings of the 2016 Conference of the North American
  Chapter of the Association for Computational Linguistics: Demonstrations},
  pages 47--51.

\bibitem[{Devlin et~al.(2018)Devlin, Chang, Lee, and
  Toutanova}]{devlin2018bert}
Jacob Devlin, Ming-Wei Chang, Kenton Lee, and Kristina Toutanova. 2018.
\newblock Bert: Pre-training of deep bidirectional transformers for language
  understanding.
\newblock \emph{arXiv preprint arXiv:1810.04805}.

\bibitem[{Eric et~al.(2019)Eric, Goel, Paul, Kumar, Sethi, Ku, Goyal, Agarwal,
  Gao, and Hakkani-Tur}]{eric2019multiwoz}
Mihail Eric, Rahul Goel, Shachi Paul, Adarsh Kumar, Abhishek Sethi, Peter Ku,
  Anuj~Kumar Goyal, Sanchit Agarwal, Shuyang Gao, and Dilek Hakkani-Tur. 2019.
\newblock Multiwoz 2.1: A consolidated multi-domain dialogue dataset with state
  corrections and state tracking baselines.
\newblock \emph{arXiv preprint arXiv:1907.01669}.

\bibitem[{Gao et~al.(2018)Gao, Galley, and Li}]{gao2018neural}
Jianfeng Gao, Michel Galley, and Lihong Li. 2018.
\newblock Neural approaches to conversational ai.
\newblock In \emph{The 41st International ACM SIGIR Conference on Research \&
  Development in Information Retrieval}, pages 1371--1374.

\bibitem[{Goo et~al.(2018)Goo, Gao, Hsu, Huo, Chen, Hsu, and
  Chen}]{goo2018slot}
Chih-Wen Goo, Guang Gao, Yun-Kai Hsu, Chih-Li Huo, Tsung-Chieh Chen, Keng-Wei
  Hsu, and Yun-Nung Chen. 2018.
\newblock Slot-gated modeling for joint slot filling and intent prediction.
\newblock In \emph{Proceedings of the 2018 Conference of the North American
  Chapter of the Association for Computational Linguistics: Human Language
  Technologies, Volume 2 (Short Papers)}, pages 753--757.

\bibitem[{Gunasekara et~al.(2020)Gunasekara, Kim, D'Haro, Rastogi, Chen, Eric,
  Hedayatnia, Gopalakrishnan, Liu, Huang et~al.}]{gunasekara2020overview}
Chulaka Gunasekara, Seokhwan Kim, Luis~Fernando D'Haro, Abhinav Rastogi,
  Yun-Nung Chen, Mihail Eric, Behnam Hedayatnia, Karthik Gopalakrishnan, Yang
  Liu, Chao-Wei Huang, et~al. 2020.
\newblock Overview of the ninth dialog system technology challenge: Dstc9.
\newblock \emph{arXiv preprint arXiv:2011.06486}.

\bibitem[{Ham et~al.(2020)Ham, Lee, Jang, and Kim}]{ham2020end}
Donghoon Ham, Jeong-Gwan Lee, Youngsoo Jang, and Kee-Eung Kim. 2020.
\newblock End-to-end neural pipeline for goal-oriented dialogue systems using
  gpt-2.
\newblock In \emph{Proceedings of the 58th Annual Meeting of the Association
  for Computational Linguistics}, pages 583--592.

\bibitem[{Hosseini-Asl et~al.(2020)Hosseini-Asl, McCann, Wu, Yavuz, and
  Socher}]{hosseini2020simple}
Ehsan Hosseini-Asl, Bryan McCann, Chien-Sheng Wu, Semih Yavuz, and Richard
  Socher. 2020.
\newblock A simple language model for task-oriented dialogue.
\newblock \emph{arXiv preprint arXiv:2005.00796}.

\bibitem[{Jhunjhunwala et~al.(2020)Jhunjhunwala, Bryant, and
  Shah}]{jhunjhunwala2020multi}
Megha Jhunjhunwala, Caleb Bryant, and Pararth Shah. 2020.
\newblock Multi-action dialog policy learning with interactive human teaching.
\newblock In \emph{Proceedings of the 21th Annual Meeting of the Special
  Interest Group on Discourse and Dialogue}, pages 290--296.

\bibitem[{Kim et~al.(2019)Kim, Galley, Gunasekara, Lee, Atkinson, Peng, Schulz,
  Gao, Li, Adada et~al.}]{kim2019eighth}
Seokhwan Kim, Michel Galley, Chulaka Gunasekara, Sungjin Lee, Adam Atkinson,
  Baolin Peng, Hannes Schulz, Jianfeng Gao, Jinchao Li, Mahmoud Adada, et~al.
  2019.
\newblock The eighth dialog system technology challenge.
\newblock \emph{arXiv preprint arXiv:1911.06394}.

\bibitem[{Kulh{\'a}nek et~al.(2021)Kulh{\'a}nek, Hude{\v{c}}ek, Nekvinda, and
  Du{\v{s}}ek}]{kulhanek2021augpt}
Jon{\'a}{\v{s}} Kulh{\'a}nek, Vojt{\v{e}}ch Hude{\v{c}}ek, Tom{\'a}{\v{s}}
  Nekvinda, and Ond{\v{r}}ej Du{\v{s}}ek. 2021.
\newblock Augpt: Dialogue with pre-trained language models and data
  augmentation.
\newblock \emph{arXiv preprint arXiv:2102.05126}.

\bibitem[{Lee et~al.(2019)Lee, Lee, and Kim}]{lee2019sumbt}
Hwaran Lee, Jinsik Lee, and Tae-Yoon Kim. 2019.
\newblock Sumbt: Slot-utterance matching for universal and scalable belief
  tracking.
\newblock \emph{arXiv preprint arXiv:1907.07421}.

\bibitem[{Lee and Stent(2016)}]{lee2016task}
Sungjin Lee and Amanda Stent. 2016.
\newblock Task lineages: Dialog state tracking for flexible interaction.
\newblock In \emph{Proceedings of the 17th Annual Meeting of the Special
  Interest Group on Discourse and Dialogue}, pages 11--21.

\bibitem[{Lei et~al.(2018)Lei, Jin, Kan, Ren, He, and Yin}]{lei2018sequicity}
Wenqiang Lei, Xisen Jin, Min-Yen Kan, Zhaochun Ren, Xiangnan He, and Dawei Yin.
  2018.
\newblock Sequicity: Simplifying task-oriented dialogue systems with single
  sequence-to-sequence architectures.
\newblock In \emph{Proceedings of the 56th Annual Meeting of the Association
  for Computational Linguistics (Volume 1: Long Papers)}, pages 1437--1447.

\bibitem[{Levin et~al.(1997)Levin, Pieraccini, and Eckert}]{levin1997learning}
Esther Levin, Roberto Pieraccini, and Wieland Eckert. 1997.
\newblock Learning dialogue strategies within the markov decision process
  framework.
\newblock In \emph{1997 IEEE Workshop on Automatic Speech Recognition and
  Understanding Proceedings}, pages 72--79. IEEE.

\bibitem[{Li and Qiu(2020)}]{li2020textat}
Linyang Li and Xipeng Qiu. 2020.
\newblock Textat: Adversarial training for natural language understanding with
  token-level perturbation.
\newblock \emph{arXiv preprint arXiv:2004.14543}.

\bibitem[{Liu and Lane(2017)}]{liu2017iterative}
Bing Liu and Ian Lane. 2017.
\newblock Iterative policy learning in end-to-end trainable task-oriented
  neural dialog models.
\newblock In \emph{2017 IEEE Automatic Speech Recognition and Understanding
  Workshop (ASRU)}, pages 482--489. IEEE.

\bibitem[{Liu et~al.(2018)Liu, Tur, Hakkani-Tur, Shah, and
  Heck}]{liu2018dialogue}
Bing Liu, Gokhan Tur, Dilek Hakkani-Tur, Pararth Shah, and Larry Heck. 2018.
\newblock Dialogue learning with human teaching and feedback in end-to-end
  trainable task-oriented dialogue systems.
\newblock \emph{arXiv preprint arXiv:1804.06512}.

\bibitem[{Liu et~al.(2020)Liu, Takanobu, Wen, Wan, Nie, Li, Li, Peng, and
  Huang}]{liu2020robustness}
Jiexi Liu, Ryuichi Takanobu, Jiaxin Wen, Dazhen Wan, Weiran Nie, Hongyan Li,
  Cheng Li, Wei Peng, and Minlie Huang. 2020.
\newblock Robustness testing of language understanding in dialog systems.
\newblock \emph{arXiv preprint arXiv:2012.15262}.

\bibitem[{Loshchilov and Hutter(2017)}]{loshchilov2017decoupled}
Ilya Loshchilov and Frank Hutter. 2017.
\newblock Decoupled weight decay regularization.
\newblock \emph{arXiv preprint arXiv:1711.05101}.

\bibitem[{Mnih et~al.(2015)Mnih, Kavukcuoglu, Silver, Rusu, Veness, Bellemare,
  Graves, Riedmiller, Fidjeland, Ostrovski et~al.}]{mnih2015human}
Volodymyr Mnih, Koray Kavukcuoglu, David Silver, Andrei~A Rusu, Joel Veness,
  Marc~G Bellemare, Alex Graves, Martin Riedmiller, Andreas~K Fidjeland, Georg
  Ostrovski, et~al. 2015.
\newblock Human-level control through deep reinforcement learning.
\newblock \emph{nature}, 518(7540):529--533.

\bibitem[{Peng et~al.(2020)Peng, Li, Li, Shayandeh, Liden, and
  Gao}]{peng2020soloist}
Baolin Peng, Chunyuan Li, Jinchao Li, Shahin Shayandeh, Lars Liden, and
  Jianfeng Gao. 2020.
\newblock Soloist: Few-shot task-oriented dialog with a single pre-trained
  auto-regressive model.
\newblock \emph{arXiv preprint arXiv:2005.05298}.

\bibitem[{Peng et~al.(2018)Peng, Li, Gao, Liu, Wong, and Su}]{peng2018deep}
Baolin Peng, Xiujun Li, Jianfeng Gao, Jingjing Liu, Kam-Fai Wong, and Shang-Yu
  Su. 2018.
\newblock Deep dyna-q: Integrating planning for task-completion dialogue policy
  learning.
\newblock \emph{arXiv preprint arXiv:1801.06176}.

\bibitem[{Peng et~al.(2017)Peng, Li, Li, Gao, Celikyilmaz, Lee, and
  Wong}]{peng2017composite}
Baolin Peng, Xiujun Li, Lihong Li, Jianfeng Gao, Asli Celikyilmaz, Sungjin Lee,
  and Kam-Fai Wong. 2017.
\newblock Composite task-completion dialogue policy learning via hierarchical
  deep reinforcement learning.
\newblock \emph{arXiv preprint arXiv:1704.03084}.

\bibitem[{Pentyala et~al.(2019)Pentyala, Liu, and Dreyer}]{pentyala2019multi}
Shiva Pentyala, Mengwen Liu, and Markus Dreyer. 2019.
\newblock Multi-task networks with universe, group, and task feature learning.
\newblock \emph{arXiv preprint arXiv:1907.01791}.

\bibitem[{Radford et~al.(2018)Radford, Narasimhan, Salimans, and
  Sutskever}]{radford2018improving}
Alec Radford, Karthik Narasimhan, Tim Salimans, and Ilya Sutskever. 2018.
\newblock Improving language understanding by generative pre-training.

\bibitem[{Radford et~al.(2019)Radford, Wu, Child, Luan, Amodei, and
  Sutskever}]{radford2019language}
Alec Radford, Jeffrey Wu, Rewon Child, David Luan, Dario Amodei, and Ilya
  Sutskever. 2019.
\newblock Language models are unsupervised multitask learners.
\newblock \emph{OpenAI blog}, 1(8):9.

\bibitem[{Sarikaya et~al.(2016)Sarikaya, Crook, Marin, Jeong, Robichaud,
  Celikyilmaz, Kim, Rochette, Khan, Liu et~al.}]{sarikaya2016overview}
Ruhi Sarikaya, Paul~A Crook, Alex Marin, Minwoo Jeong, Jean-Philippe Robichaud,
  Asli Celikyilmaz, Young-Bum Kim, Alexandre Rochette, Omar~Zia Khan, Xiaohu
  Liu, et~al. 2016.
\newblock An overview of end-to-end language understanding and dialog
  management for personal digital assistants.
\newblock In \emph{2016 ieee spoken language technology workshop (slt)}, pages
  391--397. IEEE.

\bibitem[{Schatzmann et~al.(2007)Schatzmann, Thomson, Weilhammer, Ye, and
  Young}]{schatzmann2007agenda}
Jost Schatzmann, Blaise Thomson, Karl Weilhammer, Hui Ye, and Steve Young.
  2007.
\newblock Agenda-based user simulation for bootstrapping a pomdp dialogue
  system.
\newblock In \emph{Human Language Technologies 2007: The Conference of the
  North American Chapter of the Association for Computational Linguistics;
  Companion Volume, Short Papers}, pages 149--152.

\bibitem[{Schulman et~al.(2017)Schulman, Wolski, Dhariwal, Radford, and
  Klimov}]{schulman2017proximal}
John Schulman, Filip Wolski, Prafulla Dhariwal, Alec Radford, and Oleg Klimov.
  2017.
\newblock Proximal policy optimization algorithms.
\newblock \emph{arXiv preprint arXiv:1707.06347}.

\bibitem[{Stolcke et~al.(2000)Stolcke, Ries, Coccaro, Shriberg, Bates,
  Jurafsky, Taylor, Martin, Ess-Dykema, and Meteer}]{stolcke2000dialogue}
Andreas Stolcke, Klaus Ries, Noah Coccaro, Elizabeth Shriberg, Rebecca Bates,
  Daniel Jurafsky, Paul Taylor, Rachel Martin, Carol~Van Ess-Dykema, and Marie
  Meteer. 2000.
\newblock Dialogue act modeling for automatic tagging and recognition of
  conversational speech.
\newblock \emph{Computational linguistics}, 26(3):339--373.

\bibitem[{Sutskever et~al.(2014)Sutskever, Vinyals, and
  Le}]{sutskever2014sequence}
Ilya Sutskever, Oriol Vinyals, and Quoc~V Le. 2014.
\newblock Sequence to sequence learning with neural networks.
\newblock \emph{arXiv preprint arXiv:1409.3215}.

\bibitem[{Takanobu et~al.(2019)Takanobu, Zhu, and Huang}]{takanobu2019guided}
Ryuichi Takanobu, Hanlin Zhu, and Minlie Huang. 2019.
\newblock Guided dialog policy learning: Reward estimation for multi-domain
  task-oriented dialog.
\newblock \emph{arXiv preprint arXiv:1908.10719}.

\bibitem[{Takanobu et~al.(2020)Takanobu, Zhu, Li, Peng, Gao, and
  Huang}]{takanobu2020your}
Ryuichi Takanobu, Qi~Zhu, Jinchao Li, Baolin Peng, Jianfeng Gao, and Minlie
  Huang. 2020.
\newblock Is your goal-oriented dialog model performing really well? empirical
  analysis of system-wise evaluation.
\newblock \emph{arXiv preprint arXiv:2005.07362}.

\bibitem[{Wei and Zou(2019)}]{wei2019eda}
Jason Wei and Kai Zou. 2019.
\newblock Eda: Easy data augmentation techniques for boosting performance on
  text classification tasks.
\newblock \emph{arXiv preprint arXiv:1901.11196}.

\bibitem[{Wen et~al.(2015)Wen, Gasic, Mrksic, Su, Vandyke, and
  Young}]{wen2015semantically}
Tsung-Hsien Wen, Milica Gasic, Nikola Mrksic, Pei-Hao Su, David Vandyke, and
  Steve Young. 2015.
\newblock Semantically conditioned lstm-based natural language generation for
  spoken dialogue systems.
\newblock \emph{arXiv preprint arXiv:1508.01745}.

\bibitem[{Xie et~al.(2015)Xie, Sun, Zhu, Chen, and Yu}]{xie2015recurrent}
Qizhe Xie, Kai Sun, Su~Zhu, Lu~Chen, and Kai Yu. 2015.
\newblock Recurrent polynomial network for dialogue state tracking with
  mismatched semantic parsers.
\newblock In \emph{Proceedings of the 16th Annual Meeting of the Special
  Interest Group on Discourse and Dialogue}, pages 295--304.

\bibitem[{Zhang et~al.(2021)Zhang, Lyu, Ding, Shen, Jia, Han, and
  Knight}]{zhang2021hybrid}
Boliang Zhang, Ying Lyu, Ning Ding, Tianhao Shen, Zhaoyang Jia, Kun Han, and
  Kevin Knight. 2021.
\newblock A hybrid task-oriented dialog system with domain and task adaptive
  pretraining.
\newblock \emph{arXiv preprint arXiv:2102.04506}.

\bibitem[{Zhang et~al.(2020{\natexlab{a}})Zhang, Ou, and Yu}]{zhang2020task}
Yichi Zhang, Zhijian Ou, and Zhou Yu. 2020{\natexlab{a}}.
\newblock Task-oriented dialog systems that consider multiple appropriate
  responses under the same context.
\newblock In \emph{Proceedings of the AAAI Conference on Artificial
  Intelligence}, volume~34, pages 9604--9611.

\bibitem[{Zhang et~al.(2020{\natexlab{b}})Zhang, Takanobu, Zhu, Huang, and
  Zhu}]{zhang2020recent}
Zheng Zhang, Ryuichi Takanobu, Qi~Zhu, MinLie Huang, and XiaoYan Zhu.
  2020{\natexlab{b}}.
\newblock Recent advances and challenges in task-oriented dialog systems.
\newblock \emph{Science China Technological Sciences}, pages 1--17.

\bibitem[{Zhao and Eskenazi(2016)}]{zhao2016towards}
Tiancheng Zhao and Maxine Eskenazi. 2016.
\newblock Towards end-to-end learning for dialog state tracking and management
  using deep reinforcement learning.
\newblock \emph{arXiv preprint arXiv:1606.02560}.

\bibitem[{Zhao et~al.(2019)Zhao, Xie, and Eskenazi}]{zhao2019rethinking}
Tiancheng Zhao, Kaige Xie, and Maxine Eskenazi. 2019.
\newblock Rethinking action spaces for reinforcement learning in end-to-end
  dialog agents with latent variable models.
\newblock \emph{arXiv preprint arXiv:1902.08858}.

\bibitem[{Zhu et~al.(2020)Zhu, Zhang, Fang, Li, Takanobu, Li, Peng, Gao, Zhu,
  and Huang}]{zhu2020convlab}
Qi~Zhu, Zheng Zhang, Yan Fang, Xiang Li, Ryuichi Takanobu, Jinchao Li, Baolin
  Peng, Jianfeng Gao, Xiaoyan Zhu, and Minlie Huang. 2020.
\newblock Convlab-2: An open-source toolkit for building, evaluating, and
  diagnosing dialogue systems.
\newblock \emph{arXiv preprint arXiv:2002.04793}.

\end{thebibliography}


\clearpage

\appendix

\begin{table*}[!tb]
    \centering
    \small
\begin{tabular}{c|c|c|c}
\hline
\multicolumn{2}{c|}{\textbf{Attraction}} & \multicolumn{2}{c}{\textbf{Train}} \\
\hline
Info & Reqt  & Info & Reqt  \\
\hline
area=centre & entrance fee  & arriveBy=13:00 & duration\\
type=museum &  address & day=wednesday & price\\
& &  departure=norwich &  trainID\\
& &  destination=cambridge &  \\
\hline
\end{tabular}
    \caption{User goal used in the case study. The user goal contains two domains (i.e., attraction and train). The user is supposed to inform the constraints displayed in Info, request information displayed in Reqt. The system is required to answer the request according to the user’s demands.}
    \label{case:usergoal}
\end{table*}
\begin{table*}[!tb] 
\small
\begin{tabular}{l@{~}p{7cm}|l@{~}p{7cm}}
\hline
\multicolumn{2}{c|}{\textbf{Rule}}  & \multicolumn{2}{c}{\textbf{S-PPO}} \\
\hline
U:& I am planning a trip to town and want to sight see a bit . Can you let me know some attractions I may be interested in ? A museum type of attraction . I 'd like something in the centre .  &  U:& I am planning a trip to town and want to sight see a bit . Can you let me know some attractions I may be interested in ? A museum type of attraction . I 'd like something in the centre . \\

S:& Would you like regency gallery ? Sure , there are 11 for you to choose from . & S:& They are located within the centre . We have 11 of those ! Anything specific you need or just a recommendation ? Absolutely . There are some wonderful museum in that area . You are welcome . Is there anything else I can help you with today ? \\

U:& Can I have the address of a good one ? Can you just recommend one and tell me the entrance fee ?  & U:& Just select one for me. I would like to visit on in town centre please . \\

S:& It is located in 98 king street. The entry fee is free . &  S:& How about broughton house gallery ? they 're pretty fun . That one is located in the centre . It is listed as a museum attraction . Did you need any further assistance today ? \\

U:& Yeah i need a train too. I 'd like to arrive by 13:00 , do you have something close to that time ? I would like to leave on wednesday . \textcolor{red}{Is it going to norwich ? } &  U:& Is there an exact address , like a street number ? Thanks !. Can you just recommend one and tell me the entrance fee ? \\

S:& \textcolor{red}{Where are you departing from ? } & S:& They are located at 98 king street . It is on 6 trinity street . The park is free . You are welcome . Is there anything else I can help you with today ? \\

U:& Is it going to norwich ? & U:& I also need a train. I want to get there by 13:00 . I would like to leave on wednesday. I 'll be departing from norwich . \\

S:& Where will you be leaving from ? & S:& \textcolor{blue}{Where are you headed ? When would you like the train to depart ? The arrival time is 06:35 . There are 7 . }\\

U:& \textcolor{red}{Great I also need a train departs from norwich .} & U:& Just select one for me. I don't care about the leave of the train. Ok , great thanks . Can you also help me find a train going to cambridge ? \\

S:& \textcolor{red}{I ' m sorry , we do n't have any trains arriving by 13:00 . There are no trains on wednesday . I ' m sorry there are no trains to \#TRAIN-NOOFFER-Dest\# . There are no trains leaving from norwich .} & S:& It arrives at 06:35 . TR4203 would be your perfect fit . I would say you should leave by 05:16. The price of those tickets are 17.60 pounds . \\

U:& Actually , I ' m all set . Thank you ! Bye !. &  U:& What 's the travel time on that trip ? \\

S:& Thank you for using our services . &  S:&  \textcolor{blue}{It should arrive by 06:35. The trip will cost 17.60 pounds . The trip is 79 minutes . Did you need any further assistance today ? }\\

& & U:& \textcolor{blue}{You were great . Goodbye .} \\
& & S:& It should arrive by 06:35. The fare is 17.60 pounds. I ' m happy to help , and I hope you enjoy your stay !. \\
\hline
\multicolumn{2}{c|}{\multirow[t]{2}{*}[0ex]{\textbf{Failure:}} 
\begin{tabular}[c]{@{}p{6.5cm}@{}}System does not inform the user's request. \end{tabular}} & \multicolumn{2}{c}{\textbf{Success}}
\\ \hline
\end{tabular}
\small
\centering
\caption{A sample dialog session comparison in simulated evaluation between Rule-based system and S-PPO system in natural language with the user goal in Table~\ref{case:usergoal}. 
For the rule-based system, it often provides very concise
response to users. Besides, it misunderstands the user's utterance (i.e., treat the `norwich' as destination instead of departure) which prevents it from finding satisfied entities in database. On the contrary, our S-PPO system can provide more informative sentences, understand the user's utterance correctly and thus can inform the user's request.}
\label{case:sample}
\end{table*}

\clearpage


\begin{table*}[!tb]
    \centering
    \small
\begin{tabular}{c|c|c|c}
\hline 
\multicolumn{2}{c|}{\textbf{Hotel}} & \multicolumn{2}{c}{\textbf{Train}} \\
\hline
Info & Reqt & Info & Reqt \\
\hline
parking=yes & pricerange & day=monday & duration \\
star=4  &  & depart=london liverpool street &  \\
type=guesthouse  &  & dest=cambridge  &  \\
  &  & leaveAt=15:00  &  \\
\hline
\end{tabular}
    \caption{User goal used in the case study. 
    The user goal contains two domain (i.e., hotel and train). The user is supposed to inform the constraints displayed in Info, request information displayed in Reqt. The system is required to answer the request according to the user’s demands.}
    \label{case:usergoal-humaneval}
\end{table*}
\begin{table*}[!tb] 
\small
\begin{tabular}{l@{~}p{7cm}|l@{~}p{7cm}}
\hline
\multicolumn{2}{c|}{\textbf{Rule}}  & \multicolumn{2}{c}{\textbf{S-PPO}} \\
\hline

U:& I am looking for a 4-star hotel with parking. The hotel should be in the type of guesthouse.
& 
U:& I am looking for a 4-star hotel with parking. The hotel should be in the type of guesthouse. \\

S:& Would archway house work for you ? The phone number is 01223575314 . We have 16 such places .
& 
S:& Yes , where would you like to stay and in what price range ? We have 16 moderate options , any preferences .\\

U:& What's the price range of it?
& 
U:& What's the price range of it?  \\

S:& \textcolor{red}{The aylesbray lodge guest house has free wifi . Sure , i have 16 options for you.}
& 
S:& \textcolor{blue}{It is in north and has 4 stars and is priced moderate . I have found one , Can I make a booking for you? } \\

U:& Is it expensive?
& 
U:& No, I do not need to book. Thanks! I also need a train on Monday. Depart from london liverpool street and goes to cambridge. It should leave at 15:00. \\

S:& \textcolor{red}{After further review I believe acorn guest house would be perfect for your stay !. Great , i have 16 options for you.}
& 
S:& I found one for you. The train ID is TR9202 . \\

U:& I need to know its price range.
& 
U:& How long will it take to get there? \\

S:& \textcolor{red}{That one is a cheap restaurant . It 's fantastic !. It 's in the moderate price range .}
& 
S:& \textcolor{blue}{It leaves at 15:39 and arrives at 17:07 for a total travel time of 88 minutes.} \\

U:& So is it the hotel in the cheap price range?
& 
U:& Thanks! \\
S:& \textcolor{red}{I can suggest the finches bed and breakfast . Great I have 7 different options for you !.}
& 
S:& Have a good day . Bye!\\

U:& Well, I also need a train on Monday. The train departs from london liverpool street and goes to cambridge. It should leave at 15:00.
& 
&  \\
S:& \textcolor{red}{How many tickets shall I book for you on the 15:00 train on monday from london liverpool street to cambridge ?}
& 
& \\

U:& How long will it take to get there?
& 
  \\
S:& I found a train from london liverpool street to cambridge at 15:00 on monday , will that work ?
& 
 \\

U:& How many minutes will it take to get the destination?
& 
  \\
S:& What time would you like to leave or arrive by ?
& 
 \\

\hline
\multicolumn{2}{c|}{\multirow[t]{2}{*}[0ex]{\textbf{Failure:}}
\begin{tabular}[c]{@{}p{6.5cm}@{}}System does not answer the user's requests. \end{tabular}} & \multicolumn{2}{c}{\textbf{Success}}
\\ \hline
\end{tabular}
\small
\centering
\caption{A sample dialog session comparing in human evaluation between rule-based system and S-PPO system in natural language with the user goal in Table~\ref{case:usergoal-humaneval}.
Rule-based system can not understand user's request about price range, and give contradictory responses (e.g., response `cheap' and `moderate' at the same time which confuse users). Moreover, it can not understand user's request about the duration of the train. Therefore, the dialog fails. On the contrary, S-PPO system can quickly get the point of user's request and answer user's questions correctly.
}
\label{case:sample-humaneval}
\end{table*}

\end{document}